\documentclass[conference]{IEEEtran}
\usepackage{tikz}
\usepackage{textcomp}
\usepackage{hyperref}
\usepackage{lipsum}

\newcommand\copyrighttext{%
  \footnotesize \textcopyright 2024 IEEE. Personal use of this material is permitted.
  Permission from IEEE must be obtained for all other uses, in any current or future
  media, including reprinting/republishing this material for advertising or promotional
  purposes, creating new collective works, for resale or redistribution to servers or
  lists, or reuse of any copyrighted component of this work in other works.
  DOI: \href{https://doi.org/10.1109/ICMLA61862.2024.00181}{10.1109/ICMLA61862.2024.00181}}
\newcommand\copyrightnotice{%
\begin{tikzpicture}[remember picture,overlay]
\node[anchor=south,yshift=10pt] at (current page.south) {\fbox{\parbox{\dimexpr\textwidth-\fboxsep-\fboxrule\relax}{\copyrighttext}}};
\end{tikzpicture}%
}
\usepackage{cite}
\usepackage{url}
\usepackage{hyperref}
\usepackage{amsmath,amssymb,amsfonts}
\usepackage{algorithmic}
\usepackage{graphicx}
\usepackage{caption}
\usepackage{subcaption}
\usepackage{wrapfig}
\usepackage{textcomp}
\usepackage{xcolor}
\usepackage{booktabs}
\usepackage{dcolumn}
\usepackage{tabularx}
\usepackage{multirow}
\usepackage{rotating}
\usepackage{enumitem}
\usepackage[symbol]{footmisc}

\def\BibTeX{{\rm B\kern-.05em{\sc i\kern-.025em b}\kern-.08em
    T\kern-.1667em\lower.7ex\hbox{E}\kern-.125emX}}
\begin{document}

\title{Enhancing Long-term Re-identification Robustness Using Synthetic Data: A Comparative Analysis}

\author{\IEEEauthorblockN{%
Christian Pionzewski$^{1}$,
Rebecca Rademacher$^{1}$,
Jérôme Rutinowski$^{2}$,
Antonia Ponikarov$^{1}$, \\
Stephan Matzke$^{1}$,
Tim Chilla$^{1}$,
Pia Schreynemackers$^{1}$,
Alice Kirchheim$^{1,2}$}
 
\IEEEauthorblockA{$^{1}$ Fraunhofer Institute for Material Flow and Logistics, Germany: christian.pionzewski@iml.fraunhofer.de}
\IEEEauthorblockA{$^{2}$ TU Dortmund University, Germany: jerome.rutinowski@tu-dortmund.de}

}

\maketitle
\copyrightnotice
\begin{abstract}

This contribution explores the impact of synthetic training data usage and the prediction of material wear and aging in the context of re-identification.
Different experimental setups and gallery set expanding strategies are tested, analyzing their impact on performance over time for aging re-identification subjects. 
Using a continuously updating gallery, we were able to increase our mean Rank-1 accuracy by 24\%, as material aging was taken into account step by step.
In addition, using models trained with 10\% artificial training data, Rank-1 accuracy could be increased by up to 13\%, in comparison to a model trained on only real-world data, significantly boosting generalized performance on hold-out data.
Finally, this work introduces a novel, open-source re-identification dataset, pallet-block-2696. This dataset contains 2,696 images of Euro pallets, taken over a period of 4 months.
During this time, natural aging processes occurred and some of the pallets were damaged during their usage.
These wear and tear processes significantly changed the appearance of the pallets, providing a dataset that can be used to generate synthetically aged pallets or other wooden materials. 

\end{abstract}


\section{Introduction}
\label{intro}

In many businesses and the industry as a whole, digitization is becoming a crucial factor for future success, especially for processes such as delivery and warehousing \cite{azadeh2019robotized}. 
In this context, Euro pallets and other standardized counterparts play a significant role and are often tracked and traced throughout their usage.

Previous research on this topic has already provided results on the identification of pallets in an industrial context \cite{comparison2020, rutinowski2022, schwenzfeier2023}.
The focus was on identifying pallet blocks, which are easily visible and are permanently attached to the pallet.
Since pallets block are made of chipwood, they are subject to a natural aging process that influences the visual appearance of the pallets. 
Thus, robust identification must take such aging processes into account. 
A previous approach \cite{reidrobustness2023} was to artificially expand a dataset by synthetically generating images of pallets along their life cycle and to thus simulate their aging process. 
As datasets of pallet blocks that represent realistic aging are not available, this proposal could not be validated further. 

Additionally, when employing re-identification systems that operate with changing data, it is important to consider approaches for updating the data used to re-identify industrial entities \cite{yu2023lifelong}.
Otherwise, the entities may undergo significant changes over time, rendering it impossible for the system to reliably re-identify them when relying on outdated data.
Previous work has shown that adding artificial data can be beneficial in cases where re-identification models are not exposed to entities that change over time \cite{reidrobustness2023}. 

To address these challenges, several contributions are made: A dataset of real pallet blocks, exposed to natural and realistic aging processes over a period of 4 months, is introduced to fill the gap in existing resources. In addition, three strategies for extending and updating gallery sets, which contain all known IDs, are proposed and benchmarked against this dataset to assess their effectiveness in maintaining re-identification accuracy over time. Furthermore, re-identification models from previous publications are tested on the new dataset and comparatively analyzed to validate the benefits of incorporating artificially generated training data.

The remainder of this paper is structured as follows: 
First, the related research within the area of re-identification will be discussed within Section \ref{related_literature}.
Section \ref{data_and_methods} will present the methodology and the dataset used in this contribution.
Within Section \ref{results} the key results of this work will be presented. 
Finally, Section \ref{conclusion} will provide the reader with a conclusion and an outlook of this work.

\section{Related Work}
\label{related_literature}

In the context of re-identification, a variety of models have emerged over the years, representing the state-of-the-art approaches in this relatively young research domain.
These range from well known, general-purpose deep learning models like ResNet, to task-specific ones like OSNet.
ResNet50 and other variations \cite{he2016deep}, are widely used for re-identification tasks \cite{xu2021deepchange}.
These models, which first introduced the use of residual layers, have proved to be accurate and highly accessible.
Building on the generic ResNet architecture, other approaches were developed and utilized. The PCB (Part-based Convolutional Baseline) model is one such example \cite{PCB}.
In contrast with ResNet, PCB segments the input into horizontal stripes, focusing on parts of the human body (if used on humans), e.g., head, torso, and legs.
This approach has proved to be highly efficient for pedestrian re-identification tasks, for which the model is widely used.
Another model based on residual layers, OSNet (Omni-Scale Net), is widely used as well \cite{zhou2019omni}.
In addition to the use of residual layers, OSNet uses aggregation gates that assign input-dependent and channel-wise weights to the input image's features. 
OSNet is lightweight, using pointwise and depthwise convolutions to reduce overfitting and enhance efficiency. 
Despite these lightweight characteristics, OSNet achieves state-of-the-art performance.

While such models are commonly tested on benchmark datasets like Market-1501 \cite{zheng2015scalable}, CUHK \cite{li2014deepreid}, or VIPeR \cite{gray2007evaluating}, the use of such datasets pertains to the domain of pedestrian re-identification.
For niche applications, novel datasets have to be created, as was done for animals \cite{ravoor2020deep} or industrial entities, such as pallets \cite{rutinowski2021towards, pallet-block-502, rutinowski2022}.

In this context, a first set of publications indicates potential advantages to the usage of artificial data during training to increase model accuracy, when there is a lack of data. Also, this can be beneficial in terms of model robustness (i.e., in case of visual changes occurring to an ID). 
One study demonstrated that synthetic images of pallets can be generated to augment training data without losing the inherent visual characteristics of each ID \cite{rutinowski2022applicability}.
These synthetic images retained the unique idiosyncrasies of specific pallets while varying the camera angles from which the images were captured.
Similarly, another publication \cite{reidrobustness2023} investigated the impact of synthetically aging pallet images on model robustness.
By analyzing the effect of incorporating different percentages of synthetic data into the training dataset, it was shown that models could achieve similar or improved accuracy when training with synthetic data.
Together, these studies highlight the potential of synthetic data to improve the performance of re-identification models.

\section{Data and Methods}
\label{data_and_methods}
This section outlines the methodology used in this work.
Specifically, it details the dataset creation process as well as a description of the experimental design.

\subsection{Dataset Creation Approach}
\label{subsec:datagen}
To generate the prospective dataset, we were kindly provided with 30 EPAL pallet blocks by the European Pallet Association (EPAL e. V.).
RGB images of these 30 different pallet blocks were taken from both face sides and from three perspectives (central perspective, rotated right-side, and rotated left-side), resulting in 60 unique IDs and 180 images per recording day.
The camera used was the Canon EOS 5D.
The images were taken at regular intervals of one to two weeks over a period of four months, in total at 14 different points in time, resulting in a total of 2,696.
Two sets of images were taken on the last day (later referred to as 14 and 14a), as the pallet blocks were additionally damaged (14a). 
On the first day of recording, all pallet blocks were attributed an ID and then photographed in their pristine condition.
The images were all taken at the same time of day (between 1 and 2 PM), under ambient, outdoor lighting. 
Nevertheless, variations in weather and lighting conditions resulted in variations in illumination across the recording dates.
Between the recordings, the pallet blocks remained outside, exposed to varying weather conditions.
Several damage types were simulated on the last recording day, e.g., punctual damage caused by the forks of a pallet truck and abrasive damage caused by scraping over metal, such as when a pallet passes boxes or containers.
The images were cropped to only contain the pallet block itself.  

\subsection{Experimental Design}
\label{subsec:expdesign}
In order to measure the impact of gallery updating strategies as well as test the performance of the models on the generated dataset, several series of experiments, referred to as $T$ for test, were conducted to identify the individual IDs. 
These experiments vary based on the selection of images known to the re-identification system (gallery set) and the images to be identified (query set).

When defining the image selection strategies for populating the gallery, it is crucial to weigh their advantages and disadvantages. 
A larger and more diverse set of data of the same entities in the gallery may improve matching accuracy for re-identification tasks. 
However, this increases the computational complexity of distance calculations between feature vectors, even when using indexed data structures like Faiss \cite{douze2024faiss}.\\
Another approach would be to only use the most recent image, which would have a high probability of being the most similar one and saving computational and storage overhead.
Therefore, three different series of experiments were defined. 
In the first experiment T00, images from the first two days of capture were assigned to the gallery set and the images of the other recording days were tested in the query set one day after the other. This serves as a baseline to be tested against.
Two days were used for the gallery, as the pallet blocks from the first day were in pristine condition after unpacking them, making a long-term comparison unlikely.

In the second experiment T01, rolling tests were carried out.
The gallery was successively expanded by one image set each, while the subsequent image set was used for the query test set. 
The gallery was therefore first filled with images from the first day of recording and the images from the second recording day were used to test. 
After that, the gallery was filled with the images from the first two days, and the images of the third recording day were used to test.
This principle was followed until the last day of recording, when the gallery was finally filled with the images from all 14 recording days and tested against the images from the series 14a of recordings with the damaged pallet blocks.
With this test series, it was possible to comprehend what influence the increase in information has on improving recognition and how much information provides which improvement in recognition, even after longer periods of time in comparison with less information. \\
Finally, an \emph{n+1} test was carried out in experiment T02.
Here, the recordings from only one day were stored in the gallery and tested against the recordings from the subsequent recording day.
This allows an evaluation to determine whether the strategy of successively adding more image information to the gallery has actually improved identification, compared to the strategy of including only the most recent images of a known ID.
In conclusion, the following experimental setups are given and referred to in the following sections:
\begin{itemize}
    \item T00: Experimental setup with a fixed set of gallery images that remains unchanged (recording day 01 and 02).
    \item T01: Experimental setup for which the gallery contains every image from all previous days of capture.
    \item T02: Experimental setup for which the gallery contains only images from the most recent recording day.
\end{itemize}

\subsection{Re-identification Approach}
\label{subsec:models}
The performance of pretrained state-of-the-art models will be evaluated using the herein proposed dataset.
In particular, ResNet50, PCB, and OSNet will be employed, pretrained on images of pallet blocks. 
Each model was trained either on real-world data or on a combination of real and artificial data: ResNet50$_A$, ResNet50$_R$, PCB$_A$, PCB$_R$, OSNet$_A$, and OSNet$_R$.
The designation $A$ indicates that the model was trained on a dataset consisting of a combination of $206,306$ real-world images ($62,963$ IDs) and $21,255$ artificial images ($1,635$ IDs), whereas the designation $R$ indicates that the model was exclusively trained on $330,089$ real-world images ($102,068$ IDs) \cite{{pallet-block-98382_3270},{Rutinowski:datasetB}}. 

The models are not retrained on the new dataset; rather, their pretrained performance is simply evaluated on our proposed dataset. 
The aim of this experiment is to provide insights on the suitability and generalizability of pretrained re-identification models, especially when using synthetically generated data for their training process. 

These models are evaluated using mean Average Precision and Rank-$k$ accuracy. 
The mean Average Precision (mAP) is a commonly used metric in re-identification tasks, which calculates the average precision for each ID and then averages these values. 
Rank-$k$ accuracy measures the percentage of query set instances for which a correct match is among the top $k$ candidates. 


\section{Results}
\label{results}

The results of the paper will be discussed in this section.
First, the produced dataset is described.
After that, an evaluation of the experimental setups for the gallery update strategies is given. 
Finally, a comparison of the different re-identification models and the impact of synthetic data in training is done in the last section.
The dataset, the models, as well as the code used to run the experiments, are all available online \cite{pallet-block_2696}.

\subsection{Proposed Dataset}
First, the proposed dataset and its parameters are described. 
Subsequently, a qualitative comparison of the obtained real aging data to the synthetically generated data from previous work is made. 

\subsubsection{Dataset Description}
\label{subsec:descriptiondata}
Using the methods described above, a dataset with a total of 2,696 images with a resolution of $5,472 \times 3,072$ px was created. 
After the images were created, they were manually cropped to the view of the pallet blocks without background and then scaled to $768 \times 384$ px.

\begin{figure}[h]
    \centering
    \includegraphics[width=\linewidth]{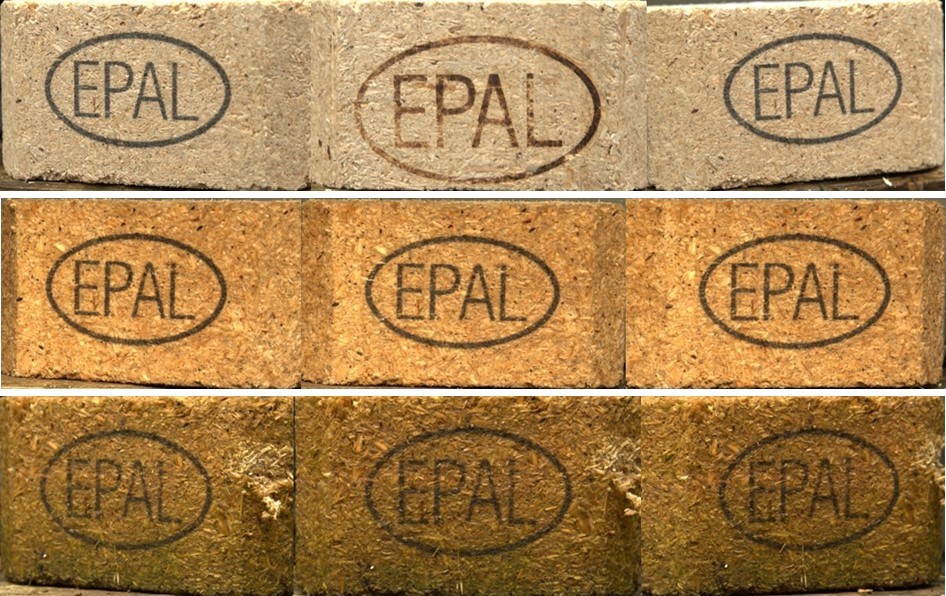} 
    \caption{Example images of ID 26. The series in each row was recorded on the following days (from top to down): recording day 1, recording day 4 and recording day 14a. Notably, the series from day 14a shows the damaged pallet blocks. }
    \label{example_images}
\end{figure}
 
Fig. \ref{example_images} shows sample images of ID 26.
These images illustrate the three different perspectives and the varying lighting conditions resulting from the differing weather conditions on the recording days. The images, from left to right, were captured from the left-hand side, from the front, and from a right-hand side perspective.
The images have been labeled with their ID, their perspective, and recording date, with additional labels indicating damage to the blocks of series 14a \cite{pallet-block_2696}.

\subsubsection{Qualitative Data Comparison}
\label{subsec:qualitativecompare}
In prior work \cite{reidrobustness2023}, synthetic pallet block images were generated in an attempt to simulate their degradation  over time.
\begin{figure}[h]
    \centering
    \includegraphics[width=\linewidth]{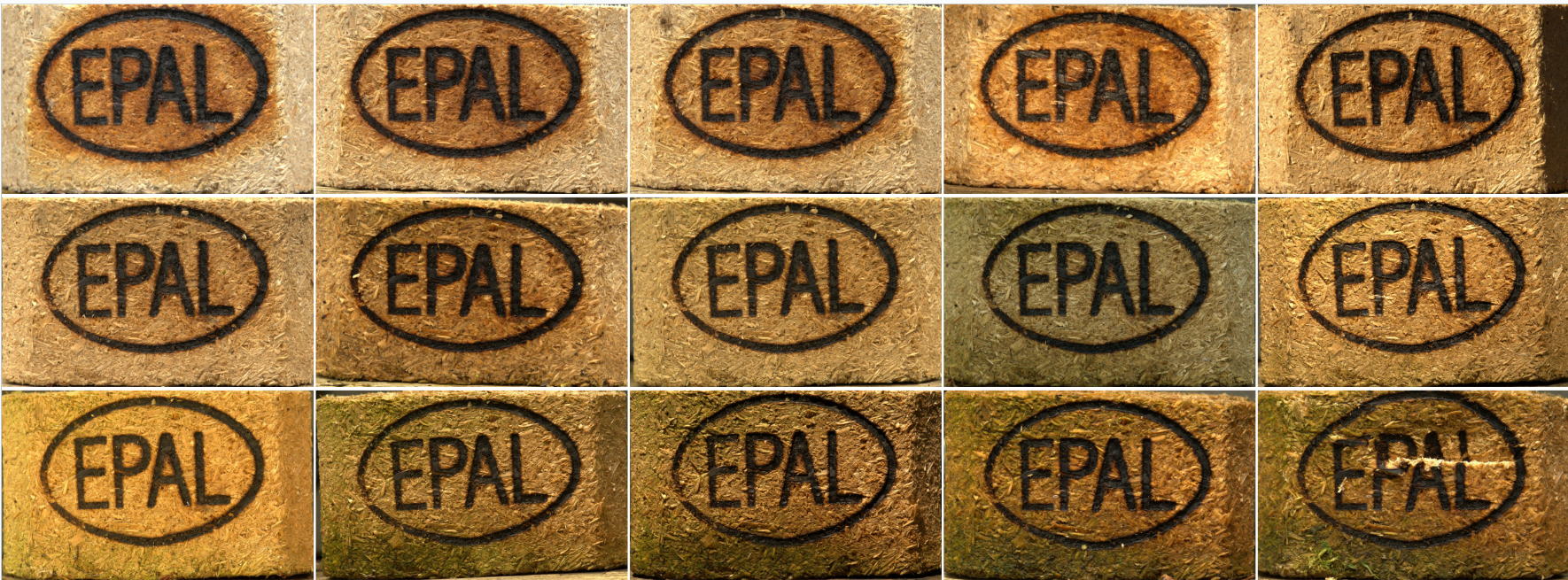} 
    \caption{The full aging process of ID 0 over four months.}
    \label{realAging_light_branding_removal_moss}
\end{figure}

While the approach helped to create more robust models for the re-identification of pallet blocks, the visual accuracy of the synthetically generated degradation was not validated.
The novel dataset of aged pallet blocks provides a new opportunity for a qualitative comparison between the simulated and real aging processes of pallets.

Fig. \ref{realAging_light_branding_removal_moss} shows the aging process of ID 0, where the top left image is the first image in the time series, while the fourth image from the left on the bottom row depicts the last. After mechanically damaging the fourth image from the left on the bottom row, the last image in the bottom right corner is obtained.

\begin{figure}[h]
    \centering
    \begin{subfigure}{\linewidth}
    \includegraphics[width=\linewidth, height=5em]{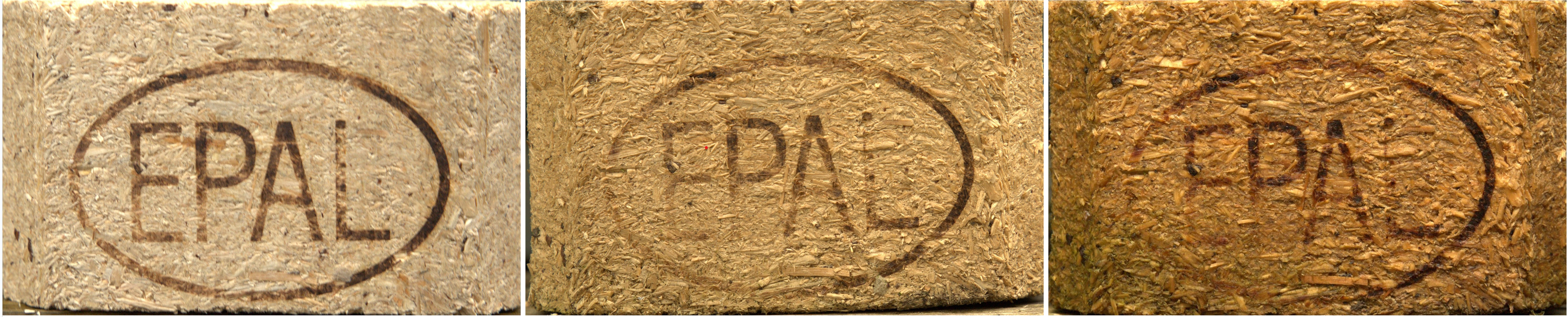} 
    \end{subfigure}
    \begin{subfigure}{\linewidth}
    \includegraphics[width=\linewidth, height=5em]{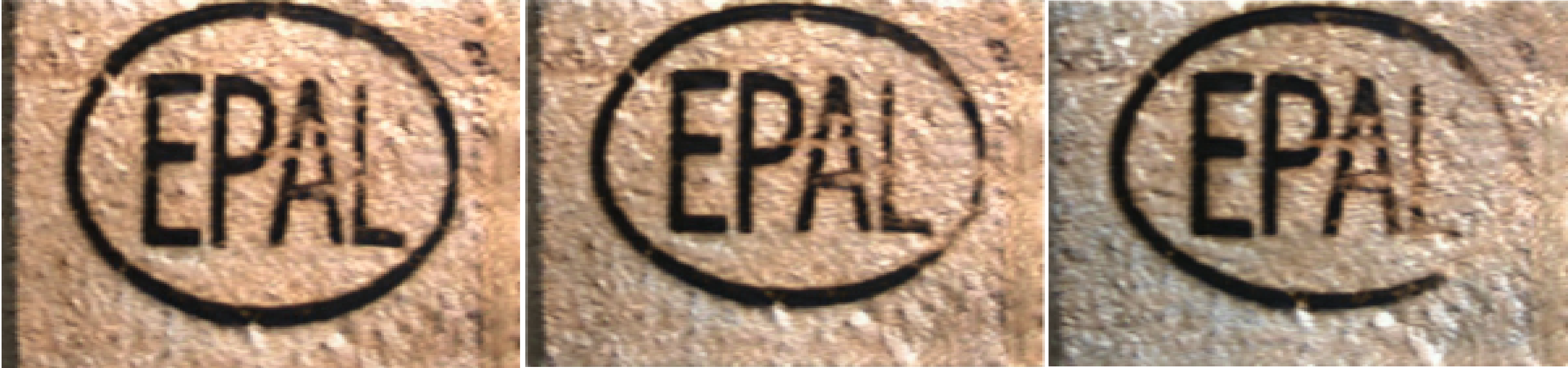} 
    \end{subfigure}
    \caption{The aging process of ID 2, the upper row showing a pronounced branding pattern decay over four months, and the lower row showing a similar aging process, generated synthetically (taken from \cite{reidrobustness2023}).}
    \label{realAging_strong_branding_removal}
\end{figure}

As can be seen in Fig. \ref{realAging_light_branding_removal_moss}, the real aging process of pallet blocks is made up of several changes. 
While the pallet block's branding initially bleeds into the areas adjacent to the branding, this bleeding effect is no longer visible after a few recording sessions.
This indicates that the branding intensity is lost over time and is further emphasized by Fig. \ref{realAging_strong_branding_removal}, where the actual branding pattern is partly lost over time. The figure further compares the synthetic aging process generated in \cite{reidrobustness2023} to the real aging process observed in this work.
The image on the left-hand side shows the original pallet block, respectively, while the middle image shows the pallet block after 2 months and the right image shows the pallet block after 4 months and their respective synthetic aging.

Another way in which pallet blocks age can be seen by the change in their surface structure.
The first few recordings of the pallet block in Fig. \ref{realAging_light_branding_removal_moss} show a clean and smooth surface structure, while the later recordings show a more distinct, worn down, and even bloated surface structure. Also, wooden materials are prone to a buildup of moss if exposed to excess humidity.
The last change pallet blocks go through while aging is not a process related to natural elements but to mechanical damage by human operators.
Looking at the last image in Fig. \ref{realAging_light_branding_removal_moss}, a horizontal cut or abrasion that stretches through most of the pallet block can be seen. 
Out of the four changes (loss of branding, change of surface structure, buildup of moss, and mechanical damage) that seem to make up pallet degradation, two were present in the synthetic data from \cite{reidrobustness2023}.
Loss of branding and change of surface structure made up large parts of the synthetic data, validating that the data synthesis process yielded realistic results.
The buildup of moss and the mechanical damage to pallet blocks were not found in the synthetic data from \cite{reidrobustness2023}. 
This can be attributed to the lack of such data in the training data for the GANs. 
In practice, the buildup of moss is a rare occurrence, as it requires leaving pallets outside on the ground, exposed to the elements for several months.

\subsection{Evaluation of Experimental Setups}
\label{subsec:reidperformance}
In this section, the results of the previously described experimental setups are laid out. 
The evaluation of the experiments for the gallery setup strategies themselves are given based on the PCB$_A$ model. 


\begin{table}[h]
    \centering
    \caption{Overview of the mean Average Precision (mAP) and Rank-$k$ accuracy for the T00 experimental setup, using the PCB$_A$ model. Query set testing starts on recording day 3, as the first two days are added to the gallery in T00.}
    \resizebox{\columnwidth}{!}{%
    \begin{tabular}{@{}rrrrrrrrr@{}}
        \toprule
        \textbf{Query Set} & \textbf{mAP} & \textbf{Rank-1} & \textbf{Rank-3} & \textbf{Rank-5} & \textbf{Rank-10} \\
        \midrule
        03 & 0.74 & 0.97 & 0.98 & 0.99 & 0.99 \\
        04 & 0.60 & 0.83 & 0.90 & 0.93 & 0.97 \\
        05 & 0.56 & 0.71 & 0.78 & 0.87 & 0.91 \\
        06 & 0.53 & 0.78 & 0.87 & 0.91 & 0.93 \\
        07 & 0.34 & 0.48 & 0.55 & 0.60 & 0.68 \\
        08 & 0.43 & 0.64 & 0.76 & 0.81 & 0.89 \\
        09 & 0.29 & 0.40 & 0.55 & 0.61 & 0.71 \\
        10 & 0.31 & 0.40 & 0.46 & 0.53 & 0.73 \\
        11 & 0.23 & 0.27 & 0.32 & 0.37 & 0.49 \\
        12 & 0.14 & 0.12 & 0.16 & 0.19 & 0.37 \\
        13 & 0.17 & 0.21 & 0.26 & 0.29 & 0.37 \\
        14 & 0.16 & 0.16 & 0.22 & 0.28 & 0.37 \\
        14a & 0.12 & 0.11 & 0.13 & 0.18 & 0.27 \\
        \midrule
        $max$ & 0.74 & 0.97 & 0.98 & 0.99 & 0.99 \\
        $min$ & 0.12 & 0.11 & 0.13 & 0.18 & 0.27 \\
        $\mu$ & 0.36 & 0.48 & 0.54 & 0.58 & 0.66 \\
        $\sigma$ & 0.21 & 0.31 & 0.31 & 0.31 & 0.27 \\
        \bottomrule
    \end{tabular}%
    }
    \label{T00pcb4r50a50metrics}
\end{table}

Tab. \ref{T00pcb4r50a50metrics} shows the results obtained by using the PCB$_A$ model that was trained on a combination of real and artificial data.
The results from the re-identification experiment indicate a decline in the performance of the model over time.
Initially, on the image set of the third recording day, the model achieves its highest mAP of 0.74 and Rank-1 accuracy of 0.97. 
With an increase in the distance between the recordings used in the gallery and the query, a significant decline in performance metrics is observed, with the mAP decreasing to 0.14 by the twelfth day of recording.
This trend suggests that the model's ability to accurately attribute IDs diminishes significantly over time. 
The damaged pallet blocks show especially low re-identification results, highlighting potential issues with the robustness of the model in handling certain degrees of damage or visual change in general. 
Overall, a mean mAP of $\mu=0.36$ with a standard deviation of $\sigma=0.21$ can be observed, indicating variability in the model's performance across recording days. 


\begin{table}[h]
    \centering
    \caption{Overview of the mean Average Precision (mAP) and Rank-$k$ accuracy for the T01 experimental setup, using the PCB$_A$ model.}
    \resizebox{\columnwidth}{!}{%
    \begin{tabular}{@{}rrrrrrrr@{}}
        \toprule
        \textbf{Query Set} & \textbf{mAP} & \textbf{Rank-1} & \textbf{Rank-3} & \textbf{Rank-5} & \textbf{Rank-10} \\
        \midrule
        02 & 0.65 & 0.77 & 0.89 & 0.93 & 0.98 \\
        03 & 0.74 & 0.97 & 0.98 & 0.99 & 0.99 \\
        04 & 0.65 & 0.87 & 0.92 & 0.94 & 0.96 \\
        05 & 0.58 & 0.76 & 0.83 & 0.87 & 0.90 \\
        06 & 0.57 & 0.89 & 0.93 & 0.94 & 0.95 \\
        07 & 0.37 & 0.68 & 0.74 & 0.78 & 0.82 \\
        08 & 0.48 & 0.82 & 0.93 & 0.96 & 0.98 \\
        09 & 0.32 & 0.82 & 0.90 & 0.96 & 0.99 \\
        10 & 0.25 & 0.54 & 0.64 & 0.68 & 0.77 \\
        11 & 0.11 & 0.46 & 0.53 & 0.61 & 0.68 \\
        12 & 0.12 & 0.41 & 0.51 & 0.58 & 0.63 \\
        13 & 0.18 & 0.63 & 0.76 & 0.79 & 0.83 \\
        14 & 0.18 & 0.70 & 0.79 & 0.81 & 0.88 \\
        14a & 0.15 & 0.75 & 0.83 & 0.86 & 0.92 \\
        \midrule
        $max$ & 0.74 & 0.97 & 0.98 & 0.99 & 0.99 \\
        $min$ & 0.11 & 0.41 & 0.51 & 0.58 & 0.63 \\
        $\mu$ & 0.39 & 0.72 & 0.79 & 0.83 & 0.87 \\
        $\sigma$ & 0.23 & 0.18 & 0.16 & 0.14 & 0.13 \\
        \bottomrule
    \end{tabular}%
    }
    \label{T01pcb4r50a50metrics}
\end{table}

The results for the T01 experiment are shown in Tab. \ref{T01pcb4r50a50metrics}.
For this experiment, a different pattern in the performance metrics can be observed.
The model starts with an mAP of 0.65 on the first query set (images from the second recording day) and achieves its peak performance on the second query set (images from the third recording day) with an mAP of 0.74 and a Rank-1 accuracy of 0.97. 
Despite the initial high performance, the metrics show a general decline over time, similar to the previous series of experiments.
A notable variance of the results per recording day is observable.
In addition, Rank-1 accuracy remains relatively high throughout the experiment, such as 0.82 on the ninth query set and 0.70 on the fourteenth query set, indicating that the expanded gallery helps maintain better identification accuracy over time. 
The mAP also fluctuates, increasing from 0.37 on the seventh query set to 0.48 on the eighth query set and still achieving 0.32 on the ninth, before eventually declining.\\
The damaged pallet blocks display a comparatively high performance in this scenario, with an mAP of 0.15 and a Rank-1 accuracy of 0.75, suggesting that the expanded gallery improves robustness in handling damaged images.
This series of experiments yields a higher mean Rank-1 accuracy of $\mu=0.72$ and a lower standard deviation of $\sigma=0.18$ than T00, suggesting more consistent re-identification performance at the first rank.
The mean mAP of $\mu=0.39$ is slightly higher than it was for the previous series of experiments, indicating better overall average performance.
The standard deviation of $\sigma=0.23$ for mAP indicates a variability that is similar to the one resulting from experiment T00.
\begin{table}[h]
    \centering
    \caption{Overview of the mean Average Precision (mAP) and Rank-$k$ accuracy for the T02 experimental setup, using the PCB$_A$ model.}
    \resizebox{\columnwidth}{!}{%
    \begin{tabular}{@{}rrrrrrrr@{}}
        \toprule
        \textbf{Query Set} & \textbf{mAP} & \textbf{Rank-1} & \textbf{Rank-3} & \textbf{Rank-5} & \textbf{Rank-10} \\
        \midrule
        02 & 0.65 & 0.77 & 0.89 & 0.93 & 0.98 \\
        03 & 0.84 & 0.96 & 0.96 & 0.97 & 0.98 \\
        04 & 0.81 & 0.84 & 0.92 & 0.94 & 0.97 \\
        05 & 0.67 & 0.69 & 0.77 & 0.81 & 0.87 \\
        06 & 0.66 & 0.78 & 0.88 & 0.89 & 0.91 \\
        07 & 0.58 & 0.61 & 0.68 & 0.72 & 0.77 \\
        08 & 0.60 & 0.58 & 0.68 & 0.79 & 0.83 \\
        09 & 0.68 & 0.71 & 0.79 & 0.85 & 0.90 \\
        10 & 0.36 & 0.37 & 0.39 & 0.44 & 0.54 \\
        11 & 0.43 & 0.46 & 0.53 & 0.61 & 0.68 \\
        12 & 0.38 & 0.36 & 0.46 & 0.52 & 0.59 \\
        13 & 0.56 & 0.61 & 0.72 & 0.76 & 0.80 \\
        14 & 0.63 & 0.68 & 0.80 & 0.84 & 0.91 \\
        14a & 0.67 & 0.76 & 0.83 & 0.85 & 0.89 \\
        \midrule
        $max$ & 0.84 & 0.96 & 0.96 & 0.97 & 0.98 \\
        $min$ & 0.36 & 0.36 & 0.39 & 0.44 & 0.54 \\
        $\mu$ & 0.61 & 0.66 & 0.73 & 0.77 & 0.82 \\
        $\sigma$ & 0.15 & 0.19 & 0.19 & 0.17 & 0.15 \\
        \bottomrule
    \end{tabular}%
    }
    \label{T02pcb4r50a50metrics}
\end{table} \\
Tab. \ref{T02pcb4r50a50metrics} shows the performance metrics on experiment T02. 
In this third and final scenario, where only the previous recording day's images are added to the gallery and tested against the current recording day, the performance metrics demonstrate a higher degree of consistency and generally increased performance, in contrast with experiment T00. 
The mAP values are also significantly higher than in for experiment T01, while slightly underperforming concerning Rank-$k$ accuracy.
The higher mAP performance was to be expected, as there are far fewer images in the gallery that are to be matched with the query set. 
Nonetheless, these findings are useful in scenarios for which more than one ID from the gallery needs to be matched or when it is challenging to retain a large amount of samples in the gallery.

\subsection{Comparison of Re-identification Model Performance}
In this section, an overall analysis of the re-identification performance of the different models over all the setups is given.

\begin{table}[h!]
    \centering
    \caption{Overview of the $\mu$ mean Average Precision (mAP) and $\mu$ Rank-$k$ accuracy for the overall experimental setups.}
    \resizebox{\columnwidth}{!}{%
    \begin{tabular}{@{}llrrrrrrr@{}}
        \toprule
        \textbf{Test} & \textbf{Model} & & \textbf{mAP} & \textbf{Rank-1} & \textbf{Rank-3} & \textbf{Rank-5} & \textbf{Rank-10} \\
        \toprule     
        \multirow{6}{*}{T00} & \multirow{2}{*}{OSNet} & \textit{A} & 0.18 & 0.23 & 0.34 & 0.41 & 0.48 \\
                             &  & \textit{R} & 0.20 & 0.25 & 0.31 & 0.35 & 0.42 \\
        \cmidrule(lr){2-8}
                             & \multirow{2}{*}{PCB} & \textit{A} & 0.36 & 0.48 & 0.54 & 0.58 & 0.66 \\
                             &  & \textit{R} & 0.33 & 0.38 & 0.45 & 0.50 & 0.57 \\
        \cmidrule(lr){2-8}
                             & \multirow{2}{*}{ResNet} & \textit{A} & 0.24 & 0.30 & 0.36 & 0.40 & 0.48 \\
                             &  & \textit{R} & 0.17 & 0.24 & 0.30 & 0.33 & 0.39 \\
        \midrule
        \multirow{6}{*}{T01} & \multirow{2}{*}{OSNet} & \textit{A} & 0.20 & 0.38 & 0.51 & 0.58 & 0.67 \\
                             &  & \textit{R} & 0.22 & 0.40 & 0.49 & 0.55 & 0.64 \\
        \cmidrule(lr){2-8}
                             & \multirow{2}{*}{PCB} & \textit{A} & 0.39 & 0.72 & 0.79 & 0.83 & 0.87 \\
                             &  & \textit{R} & 0.38 & 0.59 & 0.68 & 0.72 & 0.79 \\
        \cmidrule(lr){2-8}
                             & \multirow{2}{*}{ResNet} & \textit{A} & 0.26 & 0.45 & 0.56 & 0.62 & 0.70 \\
                             &  & \textit{R} & 0.19 & 0.36 & 0.48 & 0.55 & 0.64 \\
        \midrule
        \multirow{6}{*}{T02} & \multirow{2}{*}{OSNet} & \textit{A} & 0.35 & 0.35 & 0.46 & 0.53 & 0.62 \\
                             &  & \textit{R} & 0.36 & 0.36 & 0.45 & 0.51 & 0.58 \\
        \cmidrule(lr){2-8}
                             & \multirow{2}{*}{PCB} & \textit{A} & 0.61 & 0.66 & 0.73 & 0.77 & 0.82 \\
                             &  & \textit{R} & 0.55 & 0.54 & 0.62 & 0.66 & 0.74 \\
        \cmidrule(lr){2-8}
                             & \multirow{2}{*}{ResNet} & \textit{A} & 0.39 & 0.38 & 0.47 & 0.53 & 0.61 \\
                             &  & \textit{R} & 0.35 & 0.33 & 0.42 & 0.48 & 0.58 \\
        \bottomrule
    \end{tabular}%
    }
    \label{overallmetrics}
\end{table}
Tab. \ref{overallmetrics} compares the mean metrics of all trained re-identification models across the three experiments T00, T01, and T02. 
The mAP and Rank-k values presented here are averaged over all the mAP and Rank-k results achieved throughout the entire series of experiments.
Among the three tested neural networks, PCB overall yields the highest mAP in all three experiments.
In experiment T02, where all models achieve their best mAP results, PCB$_A$ achieves the highest performance of $\mu=0.61$ mAP when trained on artificial data.
In comparison, ResNet50$_A$ achieves $\mu=0.39$ mAP, while OSNet$_R$ achieves $\mu=0.36$ mAP. 
This indicates that the PCB model is most reliable and accurate in the task of re-identification.

During all experiments, it shows that the models that were trained on both real and artificial data consistently outperform those trained on only real data in all three experiments, with the exception of the OSNet model. 
While the OSNet$_R$ model shows slightly better results for mAP and Rank-1, both achieving $\mu=0.36$ compared to 0.35 in T02.
The models' performance on Rank-3, Rank-5, and Rank-10 was better when trained with additional artificial data. 
PCB$_A$ also achieves the highest overall accuracy for Rank-1 of $\mu=0.72$ in the experiment T01. 
In the same experiment, ResNet50$_A$ and OSNet$_R$ achieve $\mu=0.45$ and $\mu=0.40$ in Rank-1, respectively. 
This further indicates that an expanded gallery contributes to the PCB model's ability to effectively re-identify the pallet blocks. 
\section{Conclusion \& Outlook}
\label{conclusion}

This research provides an analysis of the robustness of re-identification methods in the context of aging subjects and environmental deterioration, using a novel industrial dataset, comprising $2,696$ RGB images captured at regular intervals over four months.

Several experimental setups were designed in order to measure the impact of gallery expanding strategies on re-identification performance over time, showing that regular updates to the gallery improve performance by 24\%. 

In addition, a comparison of different state-of-the-art models applied on our new dataset was made. It was shown that using models that were trained on additional synthetic data, which had the advantage of having several variations for the same IDs, outperformed models that were only trained on real-world data by up to 13\%. 
Despite these advancements, the models exhibited a certain degree of performance decline over time. 

Future research could focus on improving the robustness of the herein used models to handle various degrees of deterioration over time.
In addition, similar research could be conducted on other datasets that explicitly consider different types of variations or aging processes, to ensure the generalizability of the herein presented results.

\section{Acknowledgments} 

    This research has been funded by the Federal Ministry of Education and Research of Germany and the state of North-Rhine Westphalia as part of the Lamarr Institute for Machine Learning and Artificial Intelligence.

    This work is part of the project ``Silicon Economy Logistics Ecosystem" which is funded by the German Federal Ministry of Transport and Digital Infrastructure.
    
    We would like to thank the European Pallet Association for providing us with the pallets used for this work.

\bibliographystyle{IEEEtran}
\bibliography{IEEEabrv,citations.bib}
\end{document}